\documentclass[letterpaper, 10 pt, conference]{ieeeconf}  
\IEEEoverridecommandlockouts                              
\usepackage{url}
\usepackage{cite}
\usepackage{amsmath,amssymb,amsfonts}
\usepackage{algorithmic}
\usepackage{graphicx}
\usepackage{textcomp}
\usepackage{xcolor}
\usepackage{booktabs}
\usepackage{subcaption}
\usepackage{adjustbox}
\usepackage{pgfplots}
\usepackage{pgfplotstable}
\usepgfplotslibrary{groupplots}
\usepackage{multirow}
\usepackage{tikz}
\usetikzlibrary{spy}

\usetikzlibrary{svg.path}
\usepackage{scalerel}
\usepackage{graphicx}

\definecolor{CMT}{cmyk}{90, 0, 0, 0}
\definecolor{DeepInteraction}{cmyk}{0, 90, 0, 0}
\definecolor{TransFusion}{cmyk}{0, 90, 90, 0}
\definecolor{Sparsefusion}{cmyk}{0, 0, 0, 90}
\definecolor{BEVfusion}{cmyk}{90, 90, 0, 0}

\pgfplotsset{
    CMT_style/.style={color=CMT, solid, mark=halfcircle*, thin, mark size=1.5pt},
    DeepInteraction_style/.style={color=DeepInteraction, solid, mark=halfdiamond*, thin, mark size=1.5pt},
    TransFusion_style/.style={color=TransFusion, densely dashdotted, mark=diamond*, thin, mark size=1.5pt},
    Sparsefusion_style/.style={color=Sparsefusion, dashed, mark=star, thin, mark size=1.5pt},
    BEVfusion_style/.style={color=BEVfusion, dashdotted, mark=10-pointed star, thin, mark size=1.5pt},
}

\definecolor{orcidlogocol}{HTML}{A6CE39}
\tikzset{
  orcidlogo/.pic={
    \fill[orcidlogocol] svg{M256,128c0,70.7-57.3,128-128,128C57.3,256,0,198.7,0,128C0,57.3,57.3,0,128,0C198.7,0,256,57.3,256,128z};
    \fill[white] svg{M86.3,186.2H70.9V79.1h15.4v48.4V186.2z}
                 svg{M108.9,79.1h41.6c39.6,0,57,28.3,57,53.6c0,27.5-21.5,53.6-56.8,53.6h-41.8V79.1z M124.3,172.4h24.5c34.9,0,42.9-26.5,42.9-39.7c0-21.5-13.7-39.7-43.7-39.7h-23.7V172.4z}
                 svg{M88.7,56.8c0,5.5-4.5,10.1-10.1,10.1c-5.6,0-10.1-4.6-10.1-10.1c0-5.6,4.5-10.1,10.1-10.1C84.2,46.7,88.7,51.3,88.7,56.8z};
  }
}

\newcommand\orcidicon[1]{\href{https://orcid.org/#1}{\mbox{\scalerel*{
\begin{tikzpicture}[yscale=-1,transform shape]
\pic{orcidlogo};
\end{tikzpicture}
}{|}}}}

\usepackage{hyperref} 
    
\begin{document}

\title{MultiCorrupt: A Multi-Modal Robustness Dataset and Benchmark of LiDAR-Camera Fusion for 3D Object Detection*}

\author{Till Beemelmanns$^{1}$\textsuperscript{\orcidicon{0000-0002-2129-4082}}\,,
Quan Zhang$^{2}$\textsuperscript{\orcidicon{0000-0002-5899-1105}}\,, Christian Geller$^{1}$\textsuperscript{\orcidicon{0000-0001-8655-3201}}\,, and Lutz Eckstein$^{1}$%
\thanks{*This work has received funding from the European Union’s Horizon Europe Research and Innovation Programme under Grant Agreement No. 101076754 - AIthena project.}
\thanks{$^{1}$The authors are with the Institute for Automotive Engineering, RWTH Aachen University, 52074 Aachen, Germany {\tt\small \{firstname.lastname\}@ika.rwth-aachen.de}}%
\thanks{$^{2}$The author is with the Department of Electrical Engineering and Computer Science, TU Berlin, 10623 Berlin, Germany {\tt\small quan.zhang@campus.tu-berlin.de}}%
}






\maketitle

\begin{abstract}
Multi-modal 3D object detection models for automated driving have demonstrated exceptional performance on computer vision benchmarks like nuScenes. However, their reliance on densely sampled LiDAR point clouds and meticulously calibrated sensor arrays poses challenges for real-world applications. Issues such as sensor misalignment, miscalibration, and disparate sampling frequencies lead to spatial and temporal misalignment in data from LiDAR and cameras. Additionally, the integrity of LiDAR and camera data is often compromised by adverse environmental conditions such as inclement weather, leading to occlusions and noise interference. To address this challenge, we introduce MultiCorrupt, a comprehensive benchmark designed to evaluate the robustness of multi-modal 3D object detectors against ten distinct types of corruptions. We evaluate five state-of-the-art multi-modal detectors on MultiCorrupt and analyze their performance in terms of their resistance ability. Our results show that existing methods exhibit varying degrees of robustness depending on the type of corruption and their fusion strategy. We provide insights into which multi-modal design choices make such models robust against certain perturbations. The dataset generation code and benchmark are open-sourced at \url{https://github.com/ika-rwth-aachen/MultiCorrupt}.

\end{abstract}

\section{Introduction}

\begin{figure}[h!]
\begin{subfigure}[t]{1\linewidth}
\centering

\resizebox{\linewidth}{!}{%
\begin{tikzpicture}[spy using outlines={yellow, magnification=5, height=1.4cm, width=1.6cm, connect spies}]
\node {\pgfimage{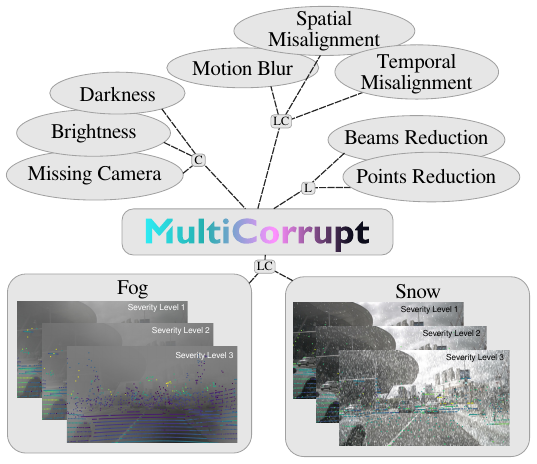}};
\spy on (2.77,-2.97) in node [right] at (0.75,-2.25);
\spy on (-1.85,-2.91) in node [right] at (-3.9,-2.25);
\end{tikzpicture}%
}%
\caption{MultiCorrupt consists of ten synthetic corruption types that affect LiDAR (L), multi-view cameras (C), or both modalities (LC).}
\label{fig:multicorrupt_overview}
\end{subfigure}%
\hfill%
\begin{subfigure}[t]{1\linewidth}
\begin{tikzpicture}
\centering 
\begin{axis}  
[  
    ybar, 
    enlargelimits=0.25,
    bar width=0.275cm,
    legend style={at={(0.50,-0.15), font=\footnotesize}, anchor=north, legend columns=3},
    y axis line style = { opacity = 0 },
    width=\linewidth,
    height=6cm,
    ylabel={nuScenes Detection Score},
    ylabel near ticks,
    symbolic x coords={Clean, Snow S2, Fog S2},  
    xtick=data,
    grid=minor,
    nodes near coords,  
    nodes near coords align={vertical}, 
    nodes near coords style={font=\scriptsize}
]  
\addplot[color=black, fill=CMT] coordinates {(Clean, 73) (Snow S2, 65) (Fog S2, 65)};
\addplot[color=black, fill=Sparsefusion] coordinates {(Clean, 73) (Snow S2, 57) (Fog S2, 63)}; 
\addplot[color=black, fill=BEVfusion] coordinates {(Clean, 71) (Snow S2, 56) (Fog S2, 60)};
\addplot[color=black, fill=TransFusion] coordinates {(Clean, 71) (Snow S2, 54) (Fog S2, 60)}; 
\addplot[color=black, fill=DeepInteraction] coordinates {(Clean, 69) (Snow S2, 53) (Fog S2, 48)}; 

\legend{CMT~\cite{yan2023crosscmt}, SparseFusion~\cite{xie2023sparsefusion}, BEVfusion\cite{mitliu2022bevfusion}, TransFusion~\cite{TransFusion} , DeepInteraction~\cite{yang2022deepinteraction}}    
\end{axis}
\end{tikzpicture}
\caption{Performance degradation of state-of-the-art multi-modal detectors for corruption \emph{Snow} and \emph{Fog} with a severity of 2. Only a subset of available corruptions is shown here.}
\label{fig:preview_benchmark}
\end{subfigure}
\caption{\textbf{MultiCorrupt: A benchmark of state-of-the-art LiDAR-camera 3D detection methods under corruption.}  (\subref{fig:multicorrupt_overview}) We introduce ten different multi-modal corruptions and (\subref{fig:preview_benchmark}) provide a comprehensive benchmark and analysis of top-performing detection models under these data perturbations.}\label{fig:overview}
\end{figure}%
Autonomous Vehicles (AVs) must comprehend their surrounding environment—vehicles, pedestrians, cyclists, and their respective postures—to further estimate the speed or future trajectories of these moving objects and plan their own motions accordingly~\cite{hendrycks2019robustness}.

While advancements have been made in the domain of autonomous driving perception, most testing and training of AVs are conducted under optimal weather and road conditions with clear visibility. Urban noise and traffic conditions significantly impact the safety and operability of AVs. For autonomous vehicles to gain widespread acceptance and adoption, they must demonstrate reliability and accuracy under all weather and road conditions. Additionally, issues like miscalibration during the vehicle's motion~\cite{limm}, and sensor misalignments~\cite{kong2023robo3d} in terms of varying frequencies or latencies, often lead to deviations between sensor modalities.

These problems are of particular interest when multi-modal detection methods are employed. Their effectiveness and robustness largely depend on \emph{how} and \emph{where} information is fused within the model. For example, early fusion combines modalities almost at the input level, making it susceptible to data corruption. On the other hand, deep fusion can be more robust, as it allows the network to learn more abstract representations, potentially mitigating the effects of information corruption or loss~\cite{jiemm}. Moreover, the way information is fused can exhibit varying degrees of sensitivity to corrupted data~\cite{limm}. 

Building on these considerations, this work introduces a consistent and open-source evaluation framework for assessing the robustness of multi-modal detection algorithms, as shown in Fig. \ref{fig:overview}. The main contributions of this paper are as follows:
\begin{itemize}
    \item We present \emph{MultiCorrupt} an open-source benchmark and dataset specifically designed for both LiDAR and image-based sensor data.
    \item We analyze five top performing multi-modal 3D object detectors on \emph{MultiCorrupt} and give valuable insights which design choice make multi-modal models robust.
\end{itemize}
We share our data generation source code and benchmark to reproduce the results presented in this work.
\section{Related Work}

\subsection{Monocular 3D Object Detection}

For monocular images, the task of detecting 3D objects inherently suffers from ambiguity. This is primarily attributed to the insufficiency of depth information that a single viewpoint can provide for accurate 3D reconstruction~\cite{chen2016monocular}. 
M3D-RPN~\cite{m3drpn} was among the first to introduce an anchor-based framework, which has since been the subject of numerous refinements, including the incorporation of differential Non-Maximum Suppression~\cite{kumar2021groomednms}, and the design of asymmetric attention modules~\cite{luo2021m3dssd}. CenterNet~\cite{duan2019centernet} was the pioneer in proposing a single-shot anchor-free framework, with subsequent research mainly focusing on its enhancement through novel depth estimation schemes~\cite{liu2020smoke,wang2022probabilisticgeodepth} or FCOS-like~\cite{tian2019fcos} architectures~\cite{wang2021fcos3d}.


\subsection{Multi-View 3D Object Detection}
AVs are commonly equipped with multiple cameras to capture a comprehensive view of the surrounding environment from various angles. 
LSS ~\cite{philion2020liftlss} serves as a seminal work to tackle the problem of BEV perception in multi-view camera setups. BEVDeT~\cite{huang2021bevdet} extends upon the LSS framework by implementing a four-stage multi-view detection pipeline to enhance its capabilities. To attain more accurate depth information, a plethora of studies have sought to extract additional information from multi-view images either through explicit depth supervision~\cite{li2023bevdepth} or stereo information~\cite{wang2022sts, li2022bevstereo}. Inspired by DETR~\cite{carion2020end}, DETR3D~\cite{wang2021detr3d} introduced transformer based detector which uses set of 3D object queries as object hypotheses. Subsequent works have refined this design, including a 3D positional embeddings on image features~\cite{liu2022petr}, and dense grid-based BEV queries~\cite{li2022bevformer}.


\subsection{Multi-Modal 3D Object Detection}
Traditionally, multi-modal fusion methods have been conveniently categorized into three paradigms: early fusion, mid fusion, and late fusion. 

\textbf{Early Fusion} approaches aim to integrate semantic information from image data into point cloud data, subsequently serving as the input for LiDAR-based 3D object methods. Frustum PointNet~\cite{qi2018frustumfPointNets} was seminal in introducing this fusion mechanism, primarily aimed at narrowing down the scope of object candidates within 3D point clouds using cues from image data. Various efforts have been made to refine Frustum PointNet; \cite{wang2019frustumConvNet} partitions the frustum into grid cells and employs CNNs over these cells, \cite{shin2018roarnet} proposes a  geometric consistency search and \cite{Frustum_PointPillars} utilizes pillar representations. Furthermore, PointPainting~\cite{vora2020pointpainting} leverages image-based semantic segmentation to enhance point clouds, and has garnered follow-up studies~\cite{simon2019complexeryolo,xu2021fusionpainting}. 

\textbf{Mid Fusion} integrates image and LiDAR features during various stages of the 3D object detection pipeline, including intermediate layers of the backbone network, proposal generation, and RoI refinement. For backbone fusion, techniques like Hybrid Voxel Feature Encoding~\cite{sindagi2019mvxnet} and Transformer approaches~\cite{zhang2022catdet,li2022deepfusion} have been utilized. Fusion methods, such as Gated Attention~\cite{3dcvf}, Unified Object Queries~\cite{chen2023futr3d}, BEV Pooling~\cite{mitliu2022bevfusion}, Learnable Alignment~\cite{chen2022autoalign,chen2022autoalignv2}, Point-to-Ray Fusion~\cite{li2022voxelfield}, and various Transformer-based techniques~\cite{TransFusion,Boost_3_D,wang2021pointaugmenting}, were applied to feature maps. Other studies explored integrating image features into point-cloud-based detection backbones~\cite{multistage,xie2019pircnn,zhu2021crossmodal}. Pioneering works like MV3D~\cite{chen2017multiviewmv3d} and AVOD~\cite{ku2018jointAVOD} use multi-view aggregation, while recent studies~\cite{chen2023futr3d,TransFusion} employ Transformer decoders for multi-modal feature fusion in RoI heads.

\textbf{Late Fusion} involves combining outputs from LiDAR-based 3D object detectors and image-based 2D object detectors, merging independently generated 2D and 3D bounding boxes for enhanced 3D detection accuracy. CLOCs~\cite{pang2020clocs} introduced a sparse tensor structure encoding paired 2D-3D boxes, refining object confidence scores. 



\subsection{Robustness of 3D Object Detection}
Object detectors face susceptibility to adversarial attacks, i.e. introducing carefully crafted perturbations to sensory inputs can mislead detection models, resulting in inaccurate or false object recognition~\cite{Sun2020TowardsRL}.

%
However, in real-world scenarios, the situation is often more complex. Due to calibration errors, misalignment, or inconsistent sampling frequencies among sensors, the data inputs are frequently subject to various degrees of bias and inaccuracies. Furthermore, portions of the data may be occluded or corrupted due to hardware and network malfunctions, environmental variables, or adverse weather conditions. These factors collectively introduce realistic challenges to the robustness of 3D object detection models~\cite{uscdmm}.%

RoboBEV~\cite{xie2023robobev} represents a comprehensive benchmark suite explicitly designed for the evaluation of \emph{multi-view camera} perception models. This suite encompasses eight distinct types of data corruption scenarios, such as \textit{Brightness}, \textit{Darkness}, and adverse weather. Extensive evaluations were conducted to investigate the resilience and reliability of these models under varying corruption conditions.%

Robo3D~\cite{kong2023robo3d} serves as a specialized evaluation suite targeted at the robustness of \emph{LiDAR-only} perception models. The authors conducted an in-depth analysis of the resilience of 3D object detectors and segmentation models for scenarios that emulate various types of real-world data corruptions. 


Yu et al. \cite{Yu2022BenchmarkingTRALI} compiled seven different LiDAR-camera noise artifacts and developed a corresponding robustness benchmark. This toolkit simulates noise conditions, such as missing camera inputs and temporal and spatial misalignment. Notably, this work does not alter the original data within the datasets; rather, it applies the perturbations to the model input during data loading, making this benchmark hard to use for a wide range of detection frameworks. In our work, we provide an easy to use dataset conversion tool which applies multiple data corruptions, including adverse weather, with several severity levels directly to the nuScenes dataset. 

Unlike specialized datasets with limited scenarios and significant domain differences \cite{bijelic2020seeingfog, Pitropov_2020cadc, DiazRuiz_2022_CVPRIthaca36}, our approach establishes a comprehensive, easy-to-use, and open-source benchmark for the robustness of LiDAR-camera fusion by synthesizing multi-modal real-world corruptions onto the widely used nuScenes dataset.


\section{Method}

\begin{figure*}[h]
  \centering
  \begin{subfigure}[b]{0.32\linewidth}
    \resizebox{\linewidth}{!}{%
    \begin{tikzpicture}[spy using outlines={yellow, magnification=2.5, height=13cm, width=16cm, connect spies}]
    \node {\pgfimage{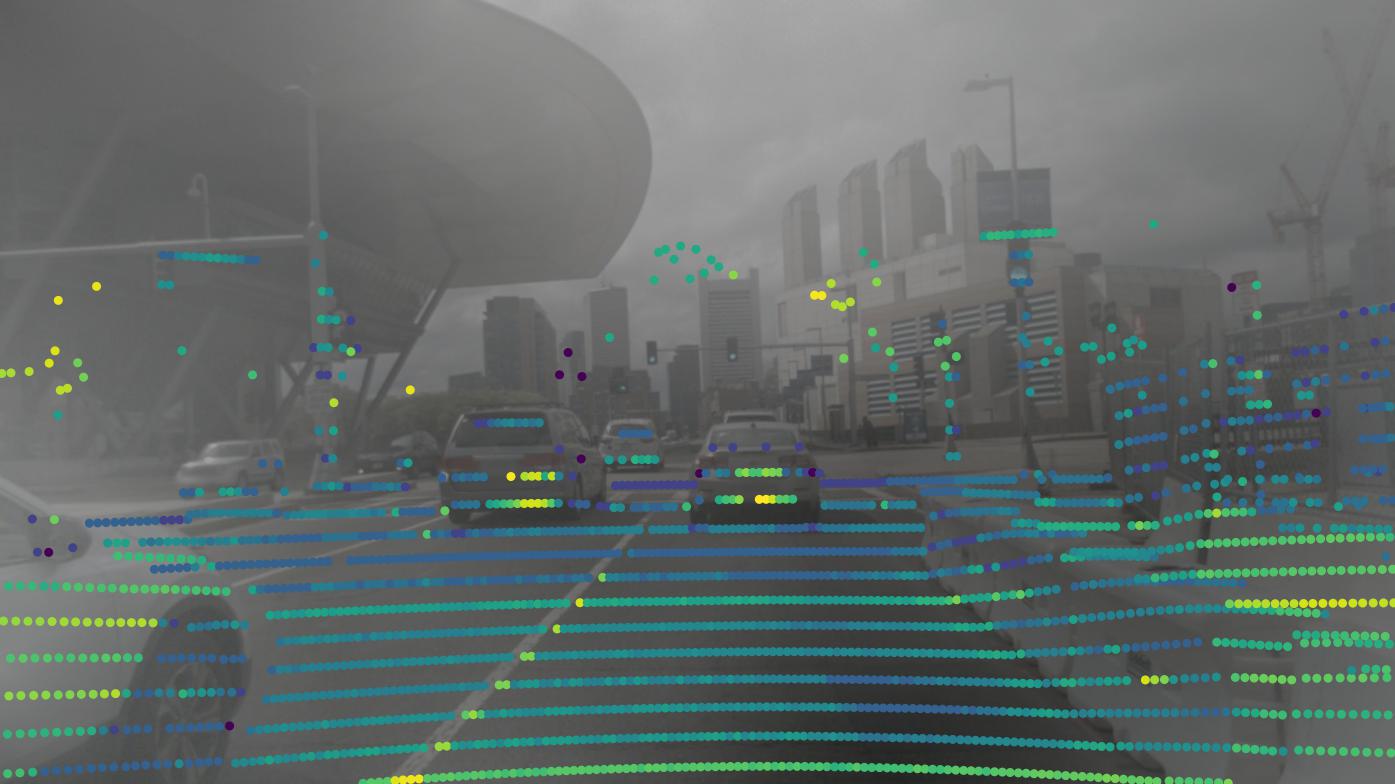}};
    \spy on (2.0,-2.75) in node [right] at (-23,6.0);
    \end{tikzpicture}%
  }%
  \caption{Fog at Severity Level 1}\label{fig:fog_severity_level_1}
  \end{subfigure}
  \begin{subfigure}[b]{0.32\linewidth}
    \resizebox{\linewidth}{!}{%
    \begin{tikzpicture}[spy using outlines={yellow, magnification=2.5, height=13cm, width=16cm, connect spies}]
    \node {\pgfimage{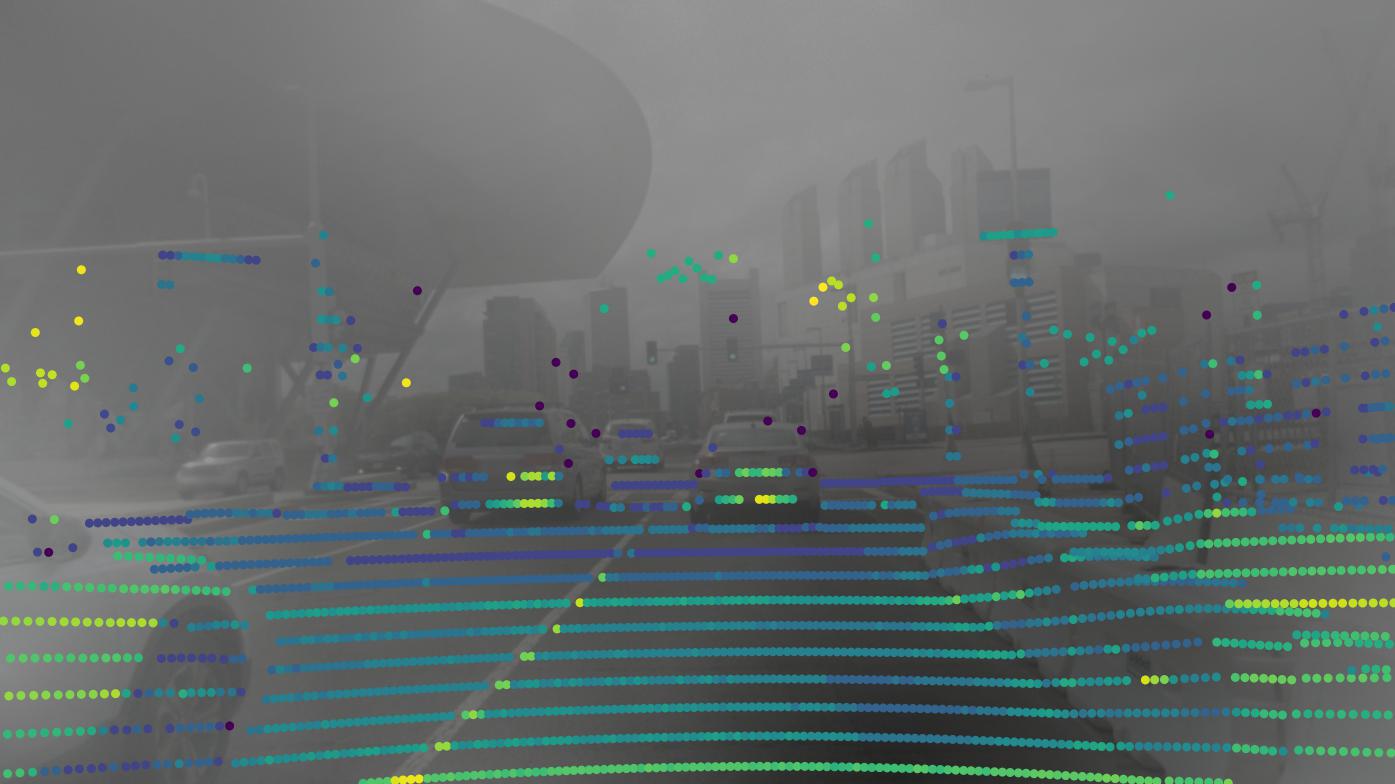}};
    \spy on (2.0,-2.75) in node [right] at (-23,6.0);
    \end{tikzpicture}%
  }%
  \caption{Fog at Severity Level 2}\label{fig:fog_severity_level_2}
  \end{subfigure}
  \begin{subfigure}[b]{0.32\linewidth}
    \resizebox{\linewidth}{!}{%
    \begin{tikzpicture}[spy using outlines={yellow, magnification=2.5, height=13cm, width=16cm, connect spies}]
    \node {\pgfimage{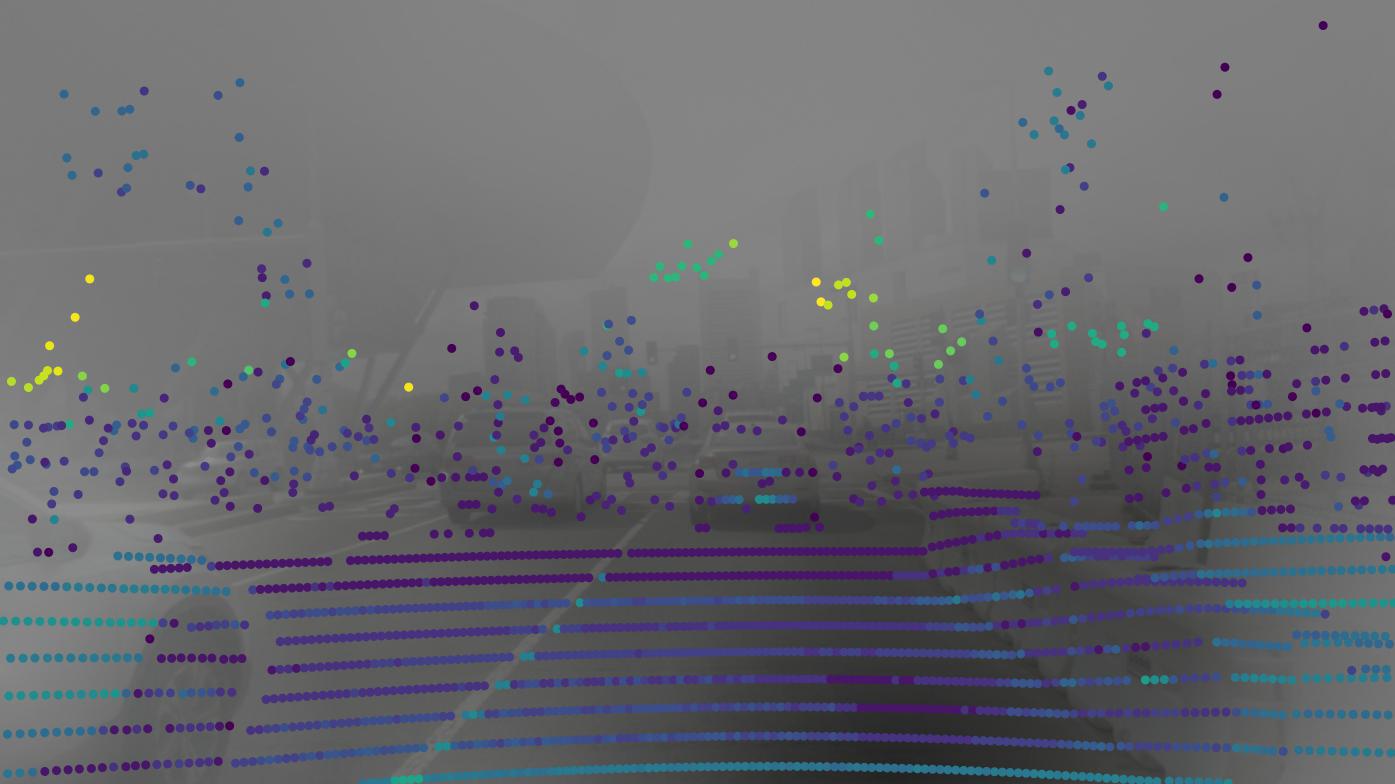}};
    \spy on (2.0,-2.75) in node [right] at (-23,6.0);
    \end{tikzpicture}%
  }%
  \caption{Fog at Severity Level 3}\label{fig:fog_severity_level_3}
  \end{subfigure}
  \begin{subfigure}[b]{0.32\linewidth}
    \resizebox{\linewidth}{!}{%
    \begin{tikzpicture}[spy using outlines={yellow, magnification=2.5, height=13cm, width=16cm, connect spies}]
    \node {\pgfimage{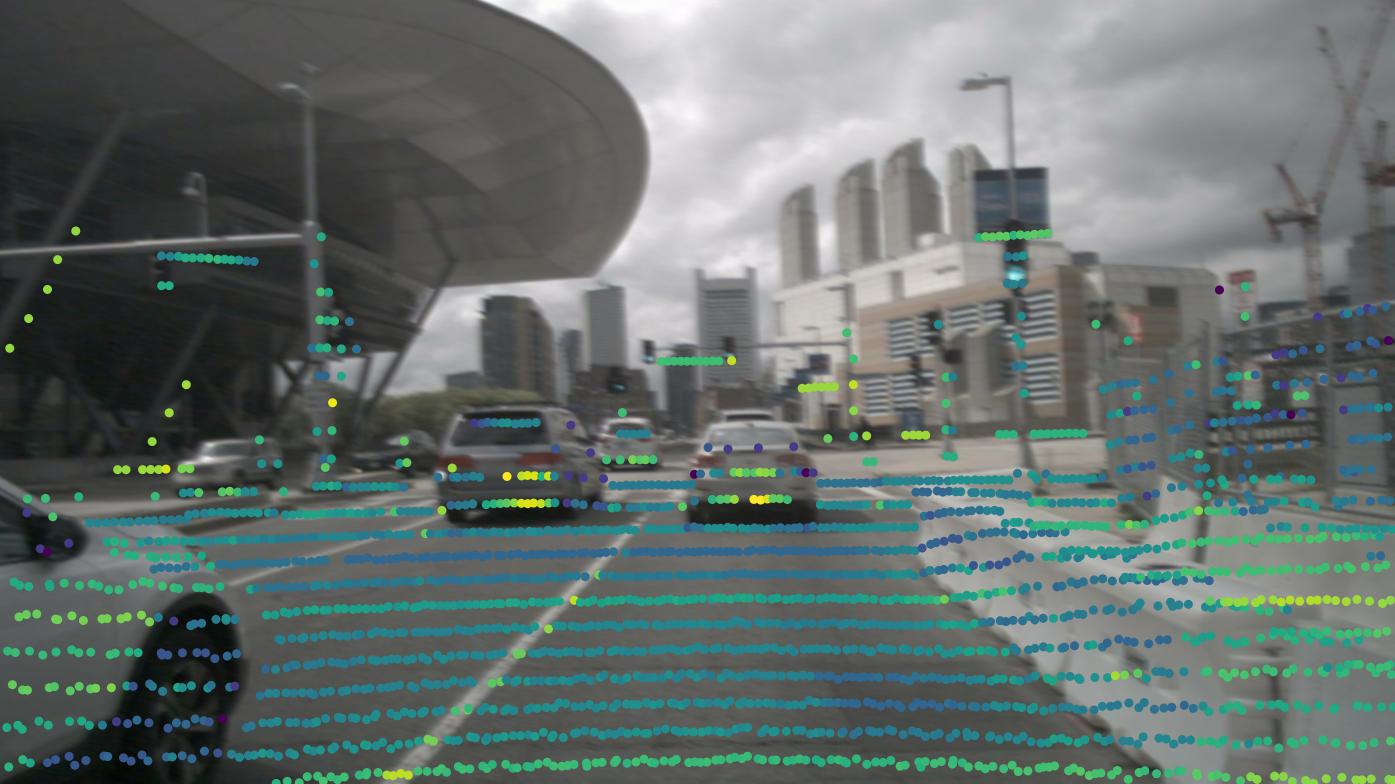}};
    \spy on (2.0,-2.75) in node [right] at (-23,6.0);
    \end{tikzpicture}%
  }%
  \caption{Motion Blur at Severity Level 1}\label{fig:motionblur_severity_level_1}
  \end{subfigure}
  \begin{subfigure}[b]{0.32\linewidth}
    \resizebox{\linewidth}{!}{%
    \begin{tikzpicture}[spy using outlines={yellow, magnification=2.5, height=13cm, width=16cm, connect spies}]
    \node {\pgfimage{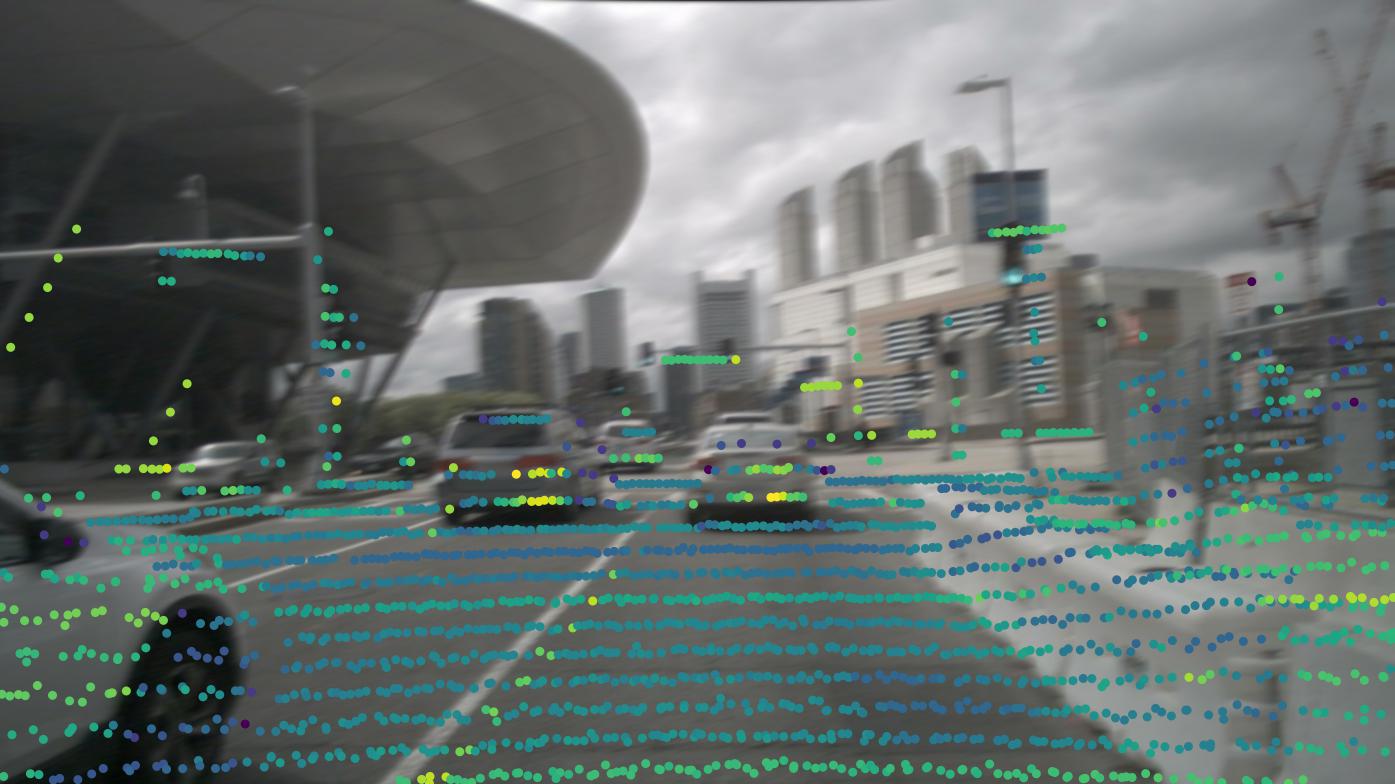}};
    \spy on (2.0,-2.75) in node [right] at (-23,6.0);
    \end{tikzpicture}%
  }%
  \caption{Motion Blur at Severity Level 2}\label{fig:motionblur_severity_level_2}
  \end{subfigure}
  \begin{subfigure}[b]{0.32\linewidth}
    \resizebox{\linewidth}{!}{%
    \begin{tikzpicture}[spy using outlines={yellow, magnification=2.5, height=13cm, width=16cm, connect spies}]
    \node {\pgfimage{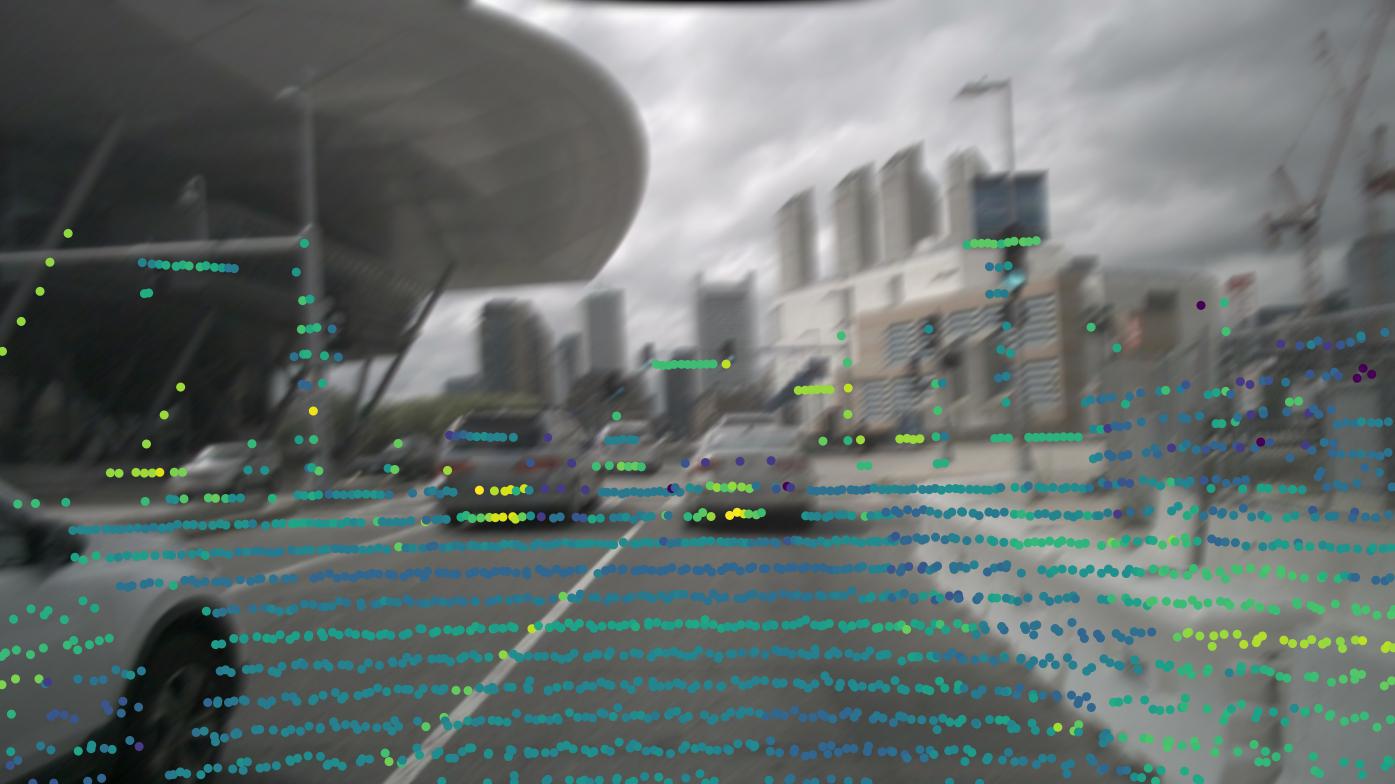}};
    \spy on (2.0,-2.75) in node [right] at (-23,6.0);
    \end{tikzpicture}%
  }%
  \caption{Motion Blur at Severity Level 3}\label{fig:motionblur_severity_level_3}
  \end{subfigure}
  \caption{\textbf{Visualization of corrupted LiDAR and camera data.} (\subref{fig:fog_severity_level_1})-(\subref{fig:fog_severity_level_3}) We display corrupted sensor data for \emph{Fog} wherein the maximum range and intensity of the LiDAR, as well as the camera image quality, degrades progressively with higher severity levels. (\subref{fig:motionblur_severity_level_1})-(\subref{fig:motionblur_severity_level_3}) The occurrence of \emph{Motion Blur} impacts both the camera and LiDAR, potentially arising from motion, vibration and the rolling shutter effect of sensors.}
  \label{fig:fog_and_motion_blur}
\end{figure*}

The success of perception systems heavily relies on their robustness and adaptability to diverse real-world conditions. In this section, we outline the comprehensive methodology employed in our study, focusing on the creation of a multi-modal corrupted dataset and the subsequent benchmark and analysis of existing multi-modal 3D object detectors.

\subsection{MultiCorrupt: Multi-Modal Corrupted Dataset}
To examine the robustness of multi-modal 3D object detectors, we introduce - \emph{MultiCorrupt} - a multi-modal corrupted dataset. This dataset undergoes deliberate corruptions, simulating challenging real-world scenarios across various environmental conditions. The corruption methods include:

\begin{itemize}
    \item \textbf{Darkness:}
    Similar to \cite{xie2023robobev}, we corrupt the multi-camera images with a Poisson Gaussian noise to emulate low light situations.
    
    \item \textbf{Brightness:}
    An overexposure corruption is introduced by adding brightness in the HSV space of the camera images.
    
    \item \textbf{Points Reducing:} We randomly drop points from the point cloud. The fraction of removed points increases with the severity level.
    
    \item \textbf{Temporal Misalignment:}
    Timestamps from different modalities, such as LiDAR and cameras, may not always be perfectly synchronized. Hence, we apply temporal misalignment to camera and point cloud data. 

    \item \textbf{Spatial Misalignment:}
    Our benchmark encompass both translation and rotation misalignment causing a spatial offset between the point cloud and camera inputs. Depending on the severity level, we vary the angle of rotation as well as the proportion of affected data within the overall dataset. 
    
    \item \textbf{Motion Blur:}
    To replicate intense motion, vibrations, and the rolling shutter effect, we incorporate jitter noise sampled from a Gaussian distribution with a standard deviation of $\sigma_t$ into the data. This noise is injected into point cloud and images, as shown in Fig. \ref{fig:fog_and_motion_blur}.
    
    \item \textbf{Missing Camera:}
    We independently and randomly drop frames from multiple cameras at each timestamp with a uniform probability, simulating the loss of some frames in a continuous time sequence.
    
    \item \textbf{Beams Reducing:} The original dataset is recorded with a 32-beam laser scanner. We  reduce the number of LiDAR beams with increasing severity levels.
    
    \item \textbf{Fog:}
    To simulate Fog in point clouds, we employ the method by Hahner et al.~\cite{HahnerCVPR22fogeth}. 
    We maintain scene consistency between images and point cloud by translating the parameters of the LiDAR fog generation into an image fog generation method, as show in Fig. \ref{fig:fog_and_motion_blur}. 
    
    \item \textbf{Snow:}
    We use an approach that samples snow particles and models them as opaque spheres~\cite{HahnerCVPR22snow}. An optical model is used to calculate the reflection properties of wet ground surfaces. We corrupt point cloud and image data corresponding to defined levels of snowfall.
\end{itemize}
A detailed overview of these methods and their specific configurations for each severity level is provided in \autoref{table:corruption_methods}.%
\begin{table*}[htbp]
    \caption{\textbf{Corruption Methods Overview}: Types, modalities, descriptions, and configurations of corruption techniques.}
    \label{table:corruption_methods}
    \centering
    \begin{tabular}{
            p{0.16\linewidth}
            p{0.05\linewidth}
            p{0.40\linewidth}
            p{0.06\linewidth}
            p{0.06\linewidth}
            p{0.06\linewidth}}
        \toprule
        \textbf{Corruption} & \textbf{Modality} & \textbf{Description} & {\textbf{Level 1}} & {\textbf{Level 2}} & {\textbf{Level 3}} \\
        \midrule
        Darkness & C & Poisson Gaussian noise intensity s & 25 & 12 & 5 \\
        Brightness & C & Addition of brightness in the HSV space & 0.5 & 0.6 & 0.7 \\
        Points Reducing & L & Dropout points with probability $p$ & 0.7 & 0.8 & 0.9 \\
        Temporal Misalignment & LC & Frozen frame applied with probability $p$ & 0.2 & 0.4 & 0.6 \\
        Spatial Misalignment & LC & Extrinsic misalignment in degrees applied with probability $p$ & 1°, 0.2 & 2°, 0.4 & 3°, 0.6 \\
        Motion Blur & LC & Jitter noise from a Gaussian distribution with $\sigma_t$ & 0.06 & 0.10 & 0.13 \\
        Missing Camera & C & Dropping frames for multiple cameras with probability $p$ & 0.2 & 0.4 & 0.6 \\
        Beams Reducing & L & Number of beams remaining in the point cloud & 16 & 8 & 4 \\
        Fog & LC & Approximated visibility in meters & 300 m & 150 m & 50 m \\
        Snow & LC & Approximated snowfall intensity in mm/h & 5 mm/h & 35 mm/h & 70 mm/h \\
        \bottomrule
    \end{tabular}
\end{table*}%
%
%
\subsection{Evaluation metrics}
We adhere to the official nuScenes metric definition~\cite{caesar2020nuScenes} for computing the NDS and mAP metrics on our \mbox{MultiCorrupt} dataset. To quantitatively compare a model's performance between the corrupted dataset and the standard nuScenes datasets, we introduce a metric called the \emph{Resistance Ability} (RA). This metric is calculated across the different severity levels with 
\begin{equation}
\text{RA}_{c,s}=\frac{\mathcal{M}_{c,s}}{\mathcal{M}_{\text{clean}}}, \text{RA}_{c}=\frac{1}{3} \sum_{s=1}^{3} \text{RA}_{c,s}, \text{mRA}=\frac{1}{N} \sum_{c=1}^{N} \text{RA}_{c}
\end{equation}
where $\mathcal{M}_{c,s}$ represents the metric for corruption $c$ at the severity level $s$,  $N$ is the total number of corruption types considered in our benchmark, and $\mathcal{M}_{\text{clean}}$ is performance on the original nuScenes dataset.

The \emph{Relative Resistance Ability} ($\mbox{RRA}_{c}$), compares the relative robustness of each model for a specific type of corruption with a baseline model. If the value is greater than zero, it indicates that the model demonstrates superior robustness compared to the baseline model. Conversely, if the value is less than zero, it suggests that the model is less robust than the baseline. We can summarize the relative resistance by computing \emph{Mean Relative Resistance Ability} ($\text{mRRA}$), which measures the relative robustness of the candidate model compared to a baseline model for all types of corruptions
\begin{equation}
\text{RRA}_{c} = \frac{\sum_{s=1}^{3} (\mathcal{M}_{c, s})}{\sum_{s=1}^{3} (\mathcal{M}_{\text{baseline}, c, s})} - 1, \label{eqmetric2a}
\end{equation}
\begin{equation}
\text{mRRA} = \frac{1}{N} \sum_{i=1}^{N} \text{RRA}_{c}. \label{eqmetric2b}
\end{equation}
$\text{RRA}_{c}$ specifically illustrates the relative robustness of each model under a particular type of corruption $c$. It allows us to scrutinize the relative performance of the models under various kinds of corruptions with the baseline. The $\text{mRRA}$ reflects the global perspective by showing the average robustness of each model across all considered types of corruption with the baseline model. 

\subsection{Benchmarking Existing Multi-Modal 3D Object Detectors}

\begin{table*}[h]
\centering
\caption{\centering{\textbf{Models under test.} Performance on nuScenes validation set and model categorization.}}
\label{table:model_comparison}
\begin{tabular}{lcccccccc}
\toprule
Method & mAP (\%) &  NDS (\%) & Representation & Alignment & Fusion Mechanism & Transformer \\
\midrule
CMT \cite{yan2023crosscmt}                    & 70.28  & 72.90 & BEV+images feature & learning \& projection & self \& cross attention & \checkmark \\ 
DeepInteraction \cite{yang2022deepinteraction}& 68.72  & 69.09  & BEV+images feature & learning \& projection & cross attention & \checkmark \\ 
TransFusion \cite{TransFusion}                & 66.72  & 70.84  & BEV+images feature & projection & image as $Q$, LiDAR as $K$ & \checkmark \\ 
Sparsefusion \cite{xie2023sparsefusion}       & 71.02  & 73.15  & BEV+images feature & learning \& projection & self-att. for LiDAR and images & \checkmark \\ 
BEVfusion \cite{mitliu2022bevfusion}          & 68.72  & 71.44  & BEV & depth and projection & concatenation & \\ 
\bottomrule
\end{tabular}
\end{table*}

We selected five top-performing multi-modal detectors from the nuScenes detection benchmark, each of which has publicly shared its model and trained weights:
\begin{itemize}
    \item \textbf{CMT} \cite{yan2023crosscmt}: Utilizes BEV and image features as represenations. Alignment is achieved through both learning and projection techniques. The fusion mechanism encompasses self-attention as well as cross-attention. CMT is trained with masked-modal training.
    \item \textbf{SparseFusion} \cite{xie2023sparsefusion}: This method employs BEV and image features for representation and uses both learning and projection for alignment. The fusion mechanism employs self-attention for both modalities. The model is conditioned to perform detection in separate LiDAR, camera and fusion branches.
    \item \textbf{TransFusion} \cite{TransFusion}: In terms of representation, this approach incorporates BEV and image features. The alignment is primarily achieved through projection. Notably, the fusion mechanism designates image features as $Q$ and LiDAR data as $K$.
    \item \textbf{DeepInteraction} \cite{yang2022deepinteraction}: Similar to CMT, this method also employs BEV and image features for representation and uses both learning and projection for alignment. It particularly emphasizes cross-attention as its fusion mechanism.
    \item \textbf{BEVfusion} \cite{mitliu2022bevfusion}: This baseline model utilizes only BEV features as representations. Alignment is achieved through depth and projection, and the fusion is achieved simply by concatenation of both modalities.
\end{itemize}
All methods employ deep fusion strategies; the primary distinctions lie in the approach how data from different modalities is fused. This typically involves three aspects:
\begin{itemize}
    \item \textbf{Representation}: Sensor data varies, leading to distinct representations such as BEV Features, Image Features, Voxels, and Range Views.
    \item \textbf{Alignment}: Alignment of modalities is generally achieved through mapping using projection matrices or through learning-based methods.
    \item \textbf{Fusion}: The most critical aspect is the fusion of data. All fusion tasks, barring BEVfusion \cite{mitliu2022bevfusion}, are accomplished through transformer-based algorithms.
\end{itemize}%
An overview of the model architectures, their performance on the nuScenes dataset, and their respective categorization is presented in \autoref{table:model_comparison}.

\section{Evaluation}
We test above listed detectors on \mbox{MultiCorrupt} and determine their RA and RRA scores, with BEVfusion \cite{mitliu2022bevfusion} chosen as the baseline model in the latter case.

\subsection{Resistance Ability \& Relative Resistance Ability}

\begin{figure*}[h]
\centering
\begin{tikzpicture}
\begin{groupplot}[
    group style={
        group size=5 by 2,
        horizontal sep=2mm,
        vertical sep=6mm,
        x descriptions at=edge bottom,
        y descriptions at=edge left
    },
    width=0.265\textwidth,
    height=3.60cm,
    xlabel={Severity Level},
    ylabel={$\mbox{RA}_{c,s}$ (NDS)},
    ylabel near ticks,
    xmin=1, xmax=3, ymin=0.4, ymax=1,
    xtick={1, 2, 3},
    ytick={0.4, 0.6, 0.8, 1},
    xlabel near ticks,
    grid style={dashed,gray!30},
    ymajorgrids=true,
    xmajorgrids=true,
    legend to name=myplotslegend,
    legend columns=-1,
    legend entries={CMT\cite{yan2023crosscmt},DeepInteraction\cite{yang2022deepinteraction},TransFusion\cite{TransFusion},SparseFusion\cite{xie2023sparsefusion},BEVfusion\cite{mitliu2022bevfusion}},
    title style={yshift=-0.23cm}
]

\nextgroupplot[title=Points Red., ymin=0.6]
\addplot[CMT_style] table[x expr=\coordindex+1, y=CMT, col sep=comma] {data/ra_results_pointsreducing.csv};
\addplot[DeepInteraction_style] table[x expr=\coordindex+1, y=DeepInteraction, col sep=comma] {data/ra_results_pointsreducing.csv};
\addplot[TransFusion_style] table[x expr=\coordindex+1, y=TransFusion, col sep=comma] {data/ra_results_pointsreducing.csv};
\addplot[Sparsefusion_style] table[x expr=\coordindex+1, y=Sparsefusion, col sep=comma] {data/ra_results_pointsreducing.csv};
\addplot[BEVfusion_style] table[x expr=\coordindex+1, y=BEVfusion, col sep=comma] {data/ra_results_pointsreducing.csv};

\nextgroupplot[title=Brightness, ymin=0.6]
\addplot[CMT_style] table[x expr=\coordindex+1, y=CMT, col sep=comma] {data/ra_results_brightness.csv};
\addplot[DeepInteraction_style] table[x expr=\coordindex+1, y=DeepInteraction, col sep=comma] {data/ra_results_brightness.csv};
\addplot[TransFusion_style] table[x expr=\coordindex+1, y=TransFusion, col sep=comma] {data/ra_results_brightness.csv};
\addplot[Sparsefusion_style] table[x expr=\coordindex+1, y=Sparsefusion, col sep=comma] {data/ra_results_brightness.csv};
\addplot[BEVfusion_style] table[x expr=\coordindex+1, y=BEVfusion, col sep=comma] {data/ra_results_brightness.csv};

\nextgroupplot[title=Darkness, ymin=0.6]
\addplot[CMT_style] table[x expr=\coordindex+1, y=CMT, col sep=comma] {data/ra_results_dark.csv};
\addplot[DeepInteraction_style] table[x expr=\coordindex+1, y=DeepInteraction, col sep=comma] {data/ra_results_dark.csv};
\addplot[TransFusion_style] table[x expr=\coordindex+1, y=TransFusion, col sep=comma] {data/ra_results_dark.csv};
\addplot[Sparsefusion_style] table[x expr=\coordindex+1, y=Sparsefusion, col sep=comma] {data/ra_results_dark.csv};
\addplot[BEVfusion_style] table[x expr=\coordindex+1, y=BEVfusion, col sep=comma] {data/ra_results_dark.csv};

\nextgroupplot[title=Motion Blur, ymin=0.6]
\addplot[CMT_style] table[x expr=\coordindex+1, y=CMT, col sep=comma] {data/ra_results_motionblur.csv};
\addplot[DeepInteraction_style] table[x expr=\coordindex+1, y=DeepInteraction, col sep=comma] {data/ra_results_motionblur.csv};
\addplot[TransFusion_style] table[x expr=\coordindex+1, y=TransFusion, col sep=comma] {data/ra_results_motionblur.csv};
\addplot[Sparsefusion_style] table[x expr=\coordindex+1, y=Sparsefusion, col sep=comma] {data/ra_results_motionblur.csv};
\addplot[BEVfusion_style] table[x expr=\coordindex+1, y=BEVfusion, col sep=comma] {data/ra_results_motionblur.csv};

\nextgroupplot[title=Missing Cam., ymin=0.6]
\addplot[CMT_style] table[x expr=\coordindex+1, y=CMT, col sep=comma] {data/ra_results_missingcamera.csv};
\addplot[DeepInteraction_style] table[x expr=\coordindex+1, y=DeepInteraction, col sep=comma] {data/ra_results_missingcamera.csv};
\addplot[TransFusion_style] table[x expr=\coordindex+1, y=TransFusion, col sep=comma] {data/ra_results_missingcamera.csv};
\addplot[Sparsefusion_style] table[x expr=\coordindex+1, y=Sparsefusion, col sep=comma] {data/ra_results_missingcamera.csv};
\addplot[BEVfusion_style] table[x expr=\coordindex+1, y=BEVfusion, col sep=comma] {data/ra_results_missingcamera.csv};

\nextgroupplot[title=Fog, ymin=0.3]
\addplot[CMT_style] table[x expr=\coordindex+1, y=CMT, col sep=comma] {data/ra_results_fog.csv};
\addplot[DeepInteraction_style] table[x expr=\coordindex+1, y=DeepInteraction, col sep=comma] {data/ra_results_fog.csv};
\addplot[TransFusion_style] table[x expr=\coordindex+1, y=TransFusion, col sep=comma] {data/ra_results_fog.csv};
\addplot[Sparsefusion_style] table[x expr=\coordindex+1, y=Sparsefusion, col sep=comma] {data/ra_results_fog.csv};
\addplot[BEVfusion_style] table[x expr=\coordindex+1, y=BEVfusion, col sep=comma] {data/ra_results_fog.csv};

\nextgroupplot[title=Beams Red., ymin=0.3]
\addplot[CMT_style] table[x expr=\coordindex+1, y=CMT, col sep=comma] {data/ra_results_beamsreducing.csv};
\addplot[DeepInteraction_style] table[x expr=\coordindex+1, y=DeepInteraction, col sep=comma] {data/ra_results_beamsreducing.csv};
\addplot[TransFusion_style] table[x expr=\coordindex+1, y=TransFusion, col sep=comma] {data/ra_results_beamsreducing.csv};
\addplot[Sparsefusion_style] table[x expr=\coordindex+1, y=Sparsefusion, col sep=comma] {data/ra_results_beamsreducing.csv};
\addplot[BEVfusion_style] table[x expr=\coordindex+1, y=BEVfusion, col sep=comma] {data/ra_results_beamsreducing.csv};

\nextgroupplot[title=Snow, ymin=0.3]
\addplot[CMT_style] table[x expr=\coordindex+1, y=CMT, col sep=comma] {data/ra_results_snow.csv};
\addplot[DeepInteraction_style] table[x expr=\coordindex+1, y=DeepInteraction, col sep=comma] {data/ra_results_snow.csv};
\addplot[TransFusion_style] table[x expr=\coordindex+1, y=TransFusion, col sep=comma] {data/ra_results_snow.csv};
\addplot[Sparsefusion_style] table[x expr=\coordindex+1, y=Sparsefusion, col sep=comma] {data/ra_results_snow.csv};
\addplot[BEVfusion_style] table[x expr=\coordindex+1, y=BEVfusion, col sep=comma] {data/ra_results_snow.csv};

\nextgroupplot[title=Spatial Mis., ymin=0.3]
\addplot[CMT_style] table[x expr=\coordindex+1, y=CMT, col sep=comma] {data/ra_results_spatialmisalignment.csv};
\addplot[DeepInteraction_style] table[x expr=\coordindex+1, y=DeepInteraction, col sep=comma] {data/ra_results_spatialmisalignment.csv};
\addplot[TransFusion_style] table[x expr=\coordindex+1, y=TransFusion, col sep=comma] {data/ra_results_spatialmisalignment.csv};
\addplot[Sparsefusion_style] table[x expr=\coordindex+1, y=Sparsefusion, col sep=comma] {data/ra_results_spatialmisalignment.csv};
\addplot[BEVfusion_style] table[x expr=\coordindex+1, y=BEVfusion, col sep=comma] {data/ra_results_spatialmisalignment.csv};

\nextgroupplot[title=Temporal Mis., ymin=0.3]
\addplot[CMT_style] table[x expr=\coordindex+1, y=CMT, col sep=comma] {data/ra_results_temporalmisalignment.csv};
\addplot[DeepInteraction_style] table[x expr=\coordindex+1, y=DeepInteraction, col sep=comma] {data/ra_results_temporalmisalignment.csv};
\addplot[TransFusion_style] table[x expr=\coordindex+1, y=TransFusion, col sep=comma] {data/ra_results_temporalmisalignment.csv};
\addplot[Sparsefusion_style] table[x expr=\coordindex+1, y=Sparsefusion, col sep=comma] {data/ra_results_temporalmisalignment.csv};
\addplot[BEVfusion_style] table[x expr=\coordindex+1, y=BEVfusion, col sep=comma] {data/ra_results_temporalmisalignment.csv};
\end{groupplot}

\node[anchor=west] at ($(group c1r2.south)!0.5!(group c5r2.south)$) [below=0.9cm] {\ref{myplotslegend}};

\end{tikzpicture}
\caption{\textbf{Robustness for all corruptions and severity levels.} {$\text{RA}_{c,s}$ for different severity levels computed using NDS score.}}
\label{fig:ndswithlevel}
\end{figure*}
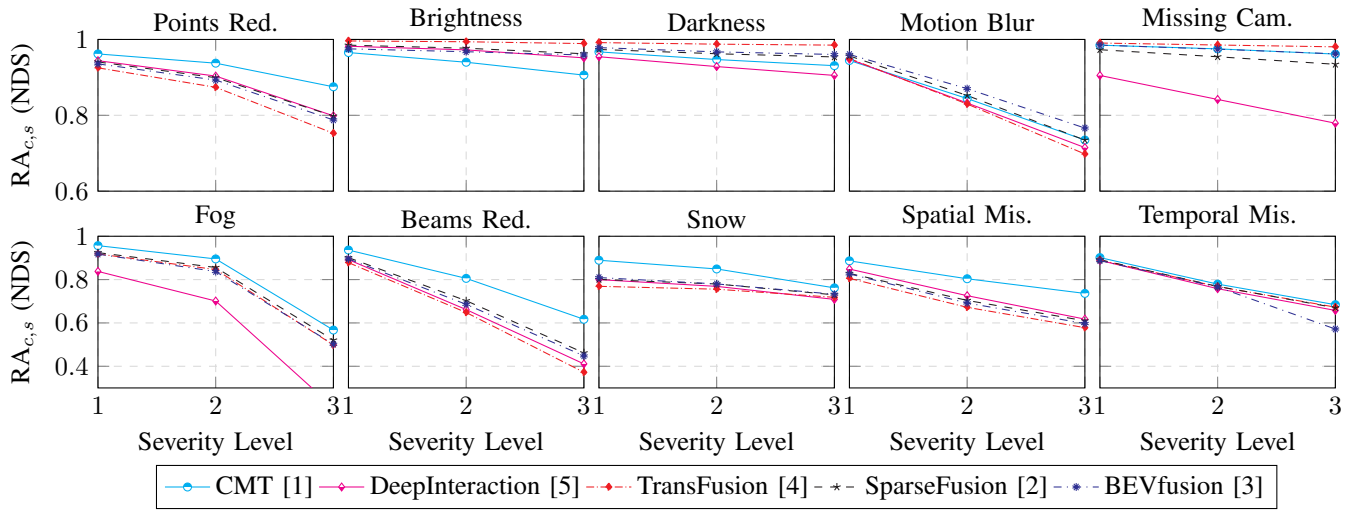

\begin{figure*}[t]
\centering
\begin{tikzpicture}
\begin{axis}[
    ybar=0pt,
    axis lines=left,
    axis x line*=middle,
    axis y line shift=0,
    width=1\textwidth,
    height=5.5cm,
    ymin=-20,
    ymax=21,
    xmin=1.75,
    bar width=0.25cm,
    enlargelimits=0.15,
    ylabel={RRA$_c$ (NDS) [\%]},
    xtick={1, 2, 3, 4, 5, 6, 7, 8, 9, 10},
    xticklabels={Beams Reduction, Brightness, Darkness, Fog, Missing Camera, Motion Blur, Points Reduction, Snow, Spatial Misalignment, Temporal Misalignment},
    xticklabel style={anchor=center, yshift=-60pt, text width=1.5cm, align=center, font=\scriptsize}, 
    major x tick style = {opacity=0},
    minor x tick num = 1,
    minor tick length=5ex,
    grid=minor,
    legend columns=-1,
    legend to name=bar_plot_legend,    
    legend entries={CMT\cite{yan2023crosscmt},DeepInteraction\cite{yang2022deepinteraction}, TransFusion\cite{TransFusion},SparseFusion\cite{xie2023sparsefusion}}
]
\addplot[color=black, fill=CMT] table [x expr=\coordindex, y=CMT, col sep=comma] {data/rra_nds_results.csv};
\addplot[color=black, fill=DeepInteraction] table [x expr=\coordindex, y=DeepInteraction, col sep=comma] {data/rra_nds_results.csv};
\addplot[color=black, fill=TransFusion] table [x expr=\coordindex, y=TransFusion, col sep=comma] {data/rra_nds_results.csv};
\addplot[color=black, fill=Sparsefusion] table [x expr=\coordindex, y=Sparsefusion, col sep=comma] {data/rra_nds_results.csv};
\end{axis}
\node[anchor=north] at (0.45\textwidth, 5.00) [below=1cm] {\ref{bar_plot_legend}};
\end{tikzpicture}
\caption{\textbf{Relative robustness visualization.} RRA$_c$ computed with NDS using BEVfusion \cite{mitliu2022bevfusion} as baseline.}
\label{fig:rraplot}
\end{figure*}
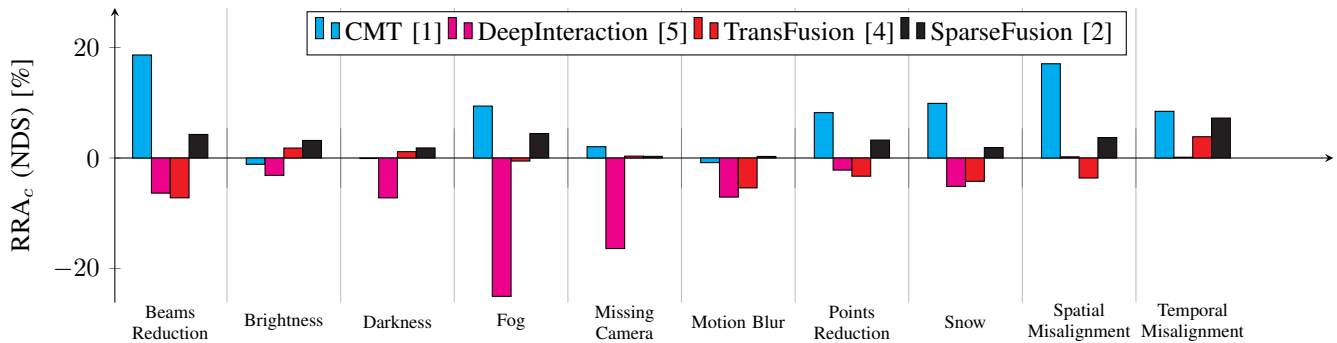

The scores $\text{RA}_c$ and mRA are shown in \autoref{tab:ra_nds}. We further visualize the impact of each severity level $\text{RA}_{c,s}$ in Fig. \ref{fig:ndswithlevel}. The consistent superior performance of CMT across all evaluation metrics is evident, making it the top-performing method. Following closely, SparseFusion emerges as the second-best performer. In contrast, both TransFusion and DeepInteraction exhibit suboptimal performance in these metrics, with their mRRA scores notably falling below the other models, as shown in  \autoref{tab:rra_nds} and Fig. \ref{fig:rraplot}.
\begin{table*}[th!]
\centering
\caption{\textbf{Robustness benchmark of state-of-the-art methods under data corruptions.} $\text{RA}_{c}$ using NDS as metric.}\label{tab:ra_nds}
\begin{adjustbox}{width=\textwidth}
\begin{tabular}{lrrrrrrrrrrr}
\toprule
Model & Beams Red. & Brightness & Darkness & Fog & Missing Cam. & Motion Blur & Points Red. & Snow & Spatial Mis. & Temporal Mis. & mRA \\
\midrule
CMT~\cite{yan2023crosscmt} & \textbf{0.786} & 0.937 & 0.948 & \textbf{0.806} & 0.974 & 0.841 & \textbf{0.925} & \textbf{0.833} & \textbf{0.809} & \textbf{0.788} & \textbf{0.865} \\
DeepInteraction~\cite{yang2022deepinteraction} & 0.655 & 0.969 & 0.929 & 0.583 & 0.842 & 0.832 & 0.882 & 0.759 & 0.731 & 0.768 & 0.795 \\
TransFusion~\cite{TransFusion} & 0.633 & \textbf{0.993} & \textbf{0.988} & 0.754 & \textbf{0.985} & 0.826 & 0.851 & 0.748 & 0.685 & 0.777 & 0.824 \\
SparseFusion~\cite{xie2023sparsefusion} & 0.689 & 0.975 & 0.963 & 0.767 & 0.954 & 0.848 & 0.879 & 0.770 & 0.714 & 0.777 & 0.834 \\
BEVfusion~\cite{mitliu2022bevfusion} & 0.676 & 0.967 & 0.969 & 0.752 & 0.974 & \textbf{0.866} & 0.872 & 0.774 & 0.705 & 0.742 & 0.830 \\
\bottomrule
\end{tabular}
\end{adjustbox}
\end{table*}
%

\begin{table*}[h]
\caption{\textbf{Relative robustness for all corruptions.} RRA$_c$ computed using NDS and BEVfusion~\cite{mitliu2022bevfusion} as baseline.}\label{tab:rra_nds}
\begin{adjustbox}{width=\textwidth}
\begin{tabular}{lrrrrrrrrrrr}
\toprule
Model & Beams Red. & Brightness & Darkness & Fog & Missing Cam. & Motion Blur & Points Red. & Snow & Spatial Mis. & Temporal Mis. & mRRA \\
\midrule
CMT~\cite{yan2023crosscmt} & \textbf{18.642} & -1.138 & -0.096 & \textbf{9.398} & \textbf{2.041} & -0.841 & \textbf{8.213} & \textbf{9.887} & \textbf{17.053} & \textbf{8.448} & \textbf{7.161} \\
DeepInteraction~\cite{yang2022deepinteraction} & -6.361 & -3.150 & -7.215 & -25.037 & -16.386 & -7.077 & -2.188 & -5.149 & 0.212 & 0.145 & -7.221 \\
TransFusion~\cite{TransFusion} & -7.210 & 1.799 & 1.146 & -0.552 & 0.340 & -5.412 & -3.296 & -4.220 & -3.626 & 3.850 & -1.718 \\
SparseFusion~\cite{xie2023sparsefusion} & 4.264 & \textbf{3.179} & \textbf{1.821} & 4.429 & 0.297 & \textbf{0.280} & 3.242 & 1.887 & 3.699 & 7.228 & 3.033 \\
\bottomrule
\end{tabular}
\end{adjustbox}
\end{table*}

\subsection{Analysis}
Despite \emph{DeepInteraction}'s reliance on multiple learning-based mechanisms, it does not consistently demonstrate robustness under various corruptions. This can be attributed to its architecture, where both the image and LiDAR modalities interact in an early stage with equal importance, making the overall performance susceptible to noise or interference from either modality. In contrast, most other methods generally employ an asymmetric fusion strategy, where at least one modality takes precedence, and the other serves to provide supplementary information. However, \emph{CMT} also treats both modalities with equal weight. Its alignment stage employs \emph{concatenation} for LiDAR and image features, resulting in a set of multi-modal tokens that are aligned with modality-specific positional embeddings. This allows for the implicit alignment of multi-modal tokens in 3D space. The queries are projected into each modality and interact with the multi-modal tokens. Hence the impact of variations in a single modality is relatively minimized as they stay independent from each other. Furthermore, \emph{CMT} is naturally trained with random missing modalities, increasing it's robustness for various corruptions. It is noteworthy that a reliance on projection matrices in the early stages of a detector, like in \emph{BEVFusion}, makes models particular sensitive to spatial misalignment. Any perturbation in the LiDAR data, such as motion blur, spatial misalignment or beams reduction, impacts \emph{TransFusion} exceptionally strong, as it's initialization of object queries relies solely on the point cloud data. \emph{SparseFusion's} robustness against many corruptions can be attributed to it's modality-specific parallel branches that basically perform detection independently from each other. Fusion is performed using sparse information exchange and a lightweight attention for final prediction. This design decision enhances robustness, ensuring that a corrupted input in one modality has a comparatively minimal impact on the other modality.
\section{Conclusion}
In this study, we introduce a robustness benchmark specifically designed for multi-modal 3D object detection. Through extensive evaluations on five top-performing multi-modal detectors, we analyze their performance in mitigating ten distinct types of data corruption. Our findings indicate that existing multi-modal 3D object detection algorithms generally exhibit different robustness behavior depending on their specific fusion, alignment and training strategies. Robustness enhancing design choices are independent modality handling, either through independent modality-spaces for Transformer tokens and queries or modality independent detection branches. Masked-modal training seems to boost robustness, but requires further analysis if it is applicable across a variety of architectures. Robustness diminishing factors are singular modality-dependant query initialization or a deep coupling of multi-modal features early in the detection pipeline. Our benchmark not only sheds light on the current landscape of robustness in multi-modal 3D object detection but also stands as a foundational tool for further investigations and advancements in the field.
%

{\small
\bibliographystyle{IEEEtran}
\bibliography{root}

\begin{thebibliography}{10}
\providecommand{\url}[1]{#1}
\csname url@samestyle\endcsname
\providecommand{\newblock}{\relax}
\providecommand{\bibinfo}[2]{#2}
\providecommand{\BIBentrySTDinterwordspacing}{\spaceskip=0pt\relax}
\providecommand{\BIBentryALTinterwordstretchfactor}{4}
\providecommand{\BIBentryALTinterwordspacing}{\spaceskip=\fontdimen2\font plus
\BIBentryALTinterwordstretchfactor\fontdimen3\font minus
  \fontdimen4\font\relax}
\providecommand{\BIBforeignlanguage}[2]{{%
\expandafter\ifx\csname l@#1\endcsname\relax
\typeout{** WARNING: IEEEtran.bst: No hyphenation pattern has been}%
\typeout{** loaded for the language `#1'. Using the pattern for}%
\typeout{** the default language instead.}%
\else
\language=\csname l@#1\endcsname
\fi
#2}}
\providecommand{\BIBdecl}{\relax}
\BIBdecl

\bibitem{yan2023crosscmt}
J.~Yan, Y.~Liu, J.~Sun, F.~Jia, S.~Li, T.~Wang, and X.~Zhang, ``Cross modal
  transformer via coordinates encoding for 3d object dectection,'' \emph{arXiv
  preprint arXiv:2301.01283}, 2023.

\bibitem{xie2023sparsefusion}
Y.~Xie, C.~Xu, M.-J. Rakotosaona, P.~Rim, F.~Tombari, K.~Keutzer, M.~Tomizuka,
  and W.~Zhan, ``Sparsefusion: Fusing multi-modal sparse representations for
  multi-sensor 3d object detection,'' \emph{arXiv preprint arXiv:2304.14340},
  2023.

\bibitem{mitliu2022bevfusion}
Z.~Liu, H.~Tang, A.~Amini, X.~Yang, H.~Mao, D.~Rus, and S.~Han, ``Bevfusion
  multi-task multi-sensor fusion with unified bird's-eye view representation,''
  2022.

\bibitem{TransFusion}
X.~Bai, Z.~Hu, X.~Zhu, Q.~Huang, Y.~Chen, H.~Fu, and C.-L. Tai, ``Transfusion:
  Robust lidar-camera fusion for 3d object detection with transformers,'' 2022.

\bibitem{yang2022deepinteraction}
Z.~Yang, J.~Chen, Z.~Miao, W.~Li, X.~Zhu, and L.~Zhang, ``Deepinteraction: 3d
  object detection via modality interaction,'' in \emph{NeurIPS}, 2022.

\bibitem{hendrycks2019robustness}
D.~Hendrycks and T.~Dietterich, ``Benchmarking neural network robustness to
  common corruptions and perturbations,'' \emph{Proceedings of the
  International Conference on Learning Representations}, 2019.

\bibitem{limm}
L.~Wang, X.~Zhang, Z.~Song, J.~Bi, G.~Zhang, H.~Wei, L.~Tang, L.~Yang, J.~Li,
  C.~Jia, and L.~Zhao, ``Multi-modal 3d object detection in autonomous driving:
  A survey and taxonomy,'' \emph{IEEE Transactions on Intelligent Vehicles},
  vol.~8, no.~7, pp. 3781--3798, 2023.

\bibitem{kong2023robo3d}
L.~Kong, Y.~Liu, X.~Li, R.~Chen, W.~Zhang, J.~Ren, L.~Pan, K.~Chen, and Z.~Liu,
  ``Robo3d: Towards robust and reliable 3d perception against corruptions,''
  \emph{arXiv preprint arXiv:2303.17597}, 2023.

\bibitem{jiemm}
Y.~Wang, Q.~Mao, H.~Zhu, J.~Deng, Y.~Zhang, J.~Ji, H.~Li, and Y.~Zhang,
  ``Multi-modal 3d object detection in autonomous driving: a survey,''
  \emph{International Journal of Computer Vision}, pp. 1--31, 2023.

\bibitem{chen2016monocular}
X.~Chen, K.~Kundu, Z.~Zhang, H.~Ma, S.~Fidler, and R.~Urtasun, ``Monocular 3d
  object detection for autonomous driving,'' in \emph{Proceedings of the IEEE
  conference on computer vision and pattern recognition}, 2016, pp. 2147--2156.

\bibitem{m3drpn}
G.~Brazil and X.~Liu, ``M3d-rpn: Monocular 3d region proposal network for
  object detection,'' in \emph{Proceedings of the IEEE International Conference
  on Computer Vision}, Seoul, South Korea, 2019.

\bibitem{kumar2021groomednms}
A.~Kumar, G.~Brazil, and X.~Liu, ``Groomed-nms: Grouped mathematically
  differentiable nms for monocular 3d object detection,'' in \emph{Proceedings
  of the IEEE/CVF conference on computer vision and pattern recognition}, 2021,
  pp. 8973--8983.

\bibitem{luo2021m3dssd}
S.~Luo, H.~Dai, L.~Shao, and Y.~Ding, ``M3dssd: Monocular 3d single stage
  object detector,'' in \emph{Proceedings of the IEEE/CVF Conference on
  Computer Vision and Pattern Recognition}, 2021, pp. 6145--6154.

\bibitem{duan2019centernet}
K.~Duan, S.~Bai, L.~Xie, H.~Qi, Q.~Huang, and Q.~Tian, ``Centernet: Keypoint
  triplets for object detection,'' 2019.

\bibitem{liu2020smoke}
Z.~Liu, Z.~Wu, and R.~T{\'o}th, ``Smoke: Single-stage monocular 3d object
  detection via keypoint estimation,'' in \emph{Proceedings of the IEEE/CVF
  Conference on Computer Vision and Pattern Recognition Workshops}, 2020, pp.
  996--997.

\bibitem{wang2022probabilisticgeodepth}
T.~Wang, Z.~Xinge, J.~Pang, and D.~Lin, ``Probabilistic and geometric depth:
  Detecting objects in perspective,'' in \emph{Conference on Robot
  Learning}.\hskip 1em plus 0.5em minus 0.4em\relax PMLR, 2022, pp. 1475--1485.

\bibitem{tian2019fcos}
Z.~Tian, C.~Shen, H.~Chen, and T.~He, ``Fcos: Fully convolutional one-stage
  object detection,'' 2019.

\bibitem{wang2021fcos3d}
T.~Wang, X.~Zhu, J.~Pang, and D.~Lin, ``Fcos3d: Fully convolutional one-stage
  monocular 3d object detection,'' in \emph{Proceedings of the IEEE/CVF
  International Conference on Computer Vision}, 2021, pp. 913--922.

\bibitem{philion2020liftlss}
J.~Philion and S.~Fidler, ``Lift, splat, shoot: Encoding images from arbitrary
  camera rigs by implicitly unprojecting to 3d,'' in \emph{Proceedings of the
  European Conference on Computer Vision}, 2020.

\bibitem{huang2021bevdet}
J.~Huang, G.~Huang, Z.~Zhu, Y.~Ye, and D.~Du, ``Bevdet: High-performance
  multi-camera 3d object detection in bird-eye-view,'' \emph{arXiv preprint
  arXiv:2112.11790}, 2021.

\bibitem{li2023bevdepth}
Y.~Li, Z.~Ge, G.~Yu, J.~Yang, Z.~Wang, Y.~Shi, J.~Sun, and Z.~Li, ``Bevdepth:
  Acquisition of reliable depth for multi-view 3d object detection,'' in
  \emph{Proceedings of the AAAI Conference on Artificial Intelligence},
  vol.~37, no.~2, 2023, pp. 1477--1485.

\bibitem{wang2022sts}
Z.~Wang, C.~Min, Z.~Ge, Y.~Li, Z.~Li, H.~Yang, and D.~Huang, ``Sts:
  Surround-view temporal stereo for multi-view 3d detection,'' \emph{arXiv
  preprint arXiv:2208.10145}, 2022.

\bibitem{li2022bevstereo}
Y.~Li, H.~Bao, Z.~Ge, J.~Yang, J.~Sun, and Z.~Li, ``Bevstereo: Enhancing depth
  estimation in multi-view 3d object detection with dynamic temporal stereo,''
  \emph{arXiv preprint arXiv:2209.10248}, 2022.

\bibitem{carion2020end}
N.~Carion, F.~Massa, G.~Synnaeve, N.~Usunier, A.~Kirillov, and S.~Zagoruyko,
  ``End-to-end object detection with transformers,'' in \emph{European
  conference on computer vision}.\hskip 1em plus 0.5em minus 0.4em\relax
  Springer, 2020, pp. 213--229.

\bibitem{wang2021detr3d}
Y.~Wang, V.~Guizilini, T.~Zhang, Y.~Wang, H.~Zhao, and J.~Solomon, ``Detr3d: 3d
  object detection from multi-view images via 3d-to-2d queries,'' 2021.

\bibitem{liu2022petr}
Y.~Liu, T.~Wang, X.~Zhang, and J.~Sun, ``Petr: Position embedding
  transformation for multi-view 3d object detection,'' in \emph{European
  Conference on Computer Vision}.\hskip 1em plus 0.5em minus 0.4em\relax
  Springer, 2022, pp. 531--548.

\bibitem{li2022bevformer}
Z.~Li, W.~Wang, H.~Li, E.~Xie, C.~Sima, T.~Lu, Q.~Yu, and J.~Dai, ``Bevformer:
  Learning bird's-eye-view representation from multi-camera images via
  spatiotemporal transformers,'' 2022.

\bibitem{qi2018frustumfPointNets}
C.~R. Qi, W.~Liu, C.~Wu, H.~Su, and L.~J. Guibas, ``Frustum pointnets for 3d
  object detection from rgb-d data,'' 2018.

\bibitem{wang2019frustumConvNet}
Z.~Wang and K.~Jia, ``Frustum convnet: Sliding frustums to aggregate local
  point-wise features for amodal 3d object detection,'' 2019.

\bibitem{shin2018roarnet}
K.~Shin, Y.~P. Kwon, and M.~Tomizuka, ``Roarnet: A robust 3d object detection
  based on region approximation refinement,'' 2018.

\bibitem{Frustum_PointPillars}
A.~Paigwar, D.~Sierra-Gonzalez, O.~Erkent, and C.~Laugier,
  ``Frustum-pointpillars: A multi-stage approach for 3d object detection using
  rgb camera and lidar,'' in \emph{2021 IEEE/CVF International Conference on
  Computer Vision Workshops (ICCVW)}, 2021, pp. 2926--2933.

\bibitem{vora2020pointpainting}
S.~Vora, A.~H. Lang, B.~Helou, and O.~Beijbom, ``Pointpainting: Sequential
  fusion for 3d object detection,'' 2020.

\bibitem{simon2019complexeryolo}
M.~Simon, K.~Amende, A.~Kraus, J.~Honer, T.~Sämann, H.~Kaulbersch, S.~Milz,
  and H.~M. Gross, ``Complexer-yolo: Real-time 3d object detection and tracking
  on semantic point clouds,'' 2019.

\bibitem{xu2021fusionpainting}
S.~Xu, D.~Zhou, J.~Fang, J.~Yin, Z.~Bin, and L.~Zhang, ``Fusionpainting:
  Multimodal fusion with adaptive attention for 3d object detection,'' 2021.

\bibitem{sindagi2019mvxnet}
V.~A. Sindagi, Y.~Zhou, and O.~Tuzel, ``Mvx-net: Multimodal voxelnet for 3d
  object detection,'' 2019.

\bibitem{zhang2022catdet}
Y.~Zhang, J.~Chen, and D.~Huang, ``Cat-det: Contrastively augmented transformer
  for multi-modal 3d object detection,'' 2022.

\bibitem{li2022deepfusion}
Y.~Li, A.~W. Yu, T.~Meng, B.~Caine, J.~Ngiam, D.~Peng, J.~Shen, B.~Wu, Y.~Lu,
  D.~Zhou, Q.~V. Le, A.~Yuille, and M.~Tan, ``Deepfusion: Lidar-camera deep
  fusion for multi-modal 3d object detection,'' 2022.

\bibitem{3dcvf}
J.~H. Yoo, Y.~Kim, J.~Kim, and J.~W. Choi, ``3d-{CVF}: Generating joint camera
  and {LiDAR} features using cross-view spatial feature fusion for 3d object
  detection,'' in \emph{Computer Vision {\textendash} {ECCV} 2020}.\hskip 1em
  plus 0.5em minus 0.4em\relax Springer International Publishing, 2020, pp.
  720--736.

\bibitem{chen2023futr3d}
X.~Chen, T.~Zhang, Y.~Wang, Y.~Wang, and H.~Zhao, ``Futr3d: A unified sensor
  fusion framework for 3d detection,'' 2023.

\bibitem{chen2022autoalign}
Z.~Chen, Z.~Li, S.~Zhang, L.~Fang, Q.~Jiang, F.~Zhao, B.~Zhou, and H.~Zhao,
  ``Autoalign: Pixel-instance feature aggregation for multi-modal 3d object
  detection,'' 2022.

\bibitem{chen2022autoalignv2}
Z.~Chen, Z.~Li, S.~Zhang, L.~Fang, Q.~Jiang, and F.~Zhao, ``Autoalignv2:
  Deformable feature aggregation for dynamic multi-modal 3d object detection,''
  \emph{ECCV}, 2022.

\bibitem{li2022voxelfield}
Y.~Li, X.~Qi, Y.~Chen, L.~Wang, Z.~Li, J.~Sun, and J.~Jia, ``Voxel field fusion
  for 3d object detection,'' 2022.

\bibitem{Boost_3_D}
Y.~Chen, H.~Li, R.~Gao, and D.~Zhao, ``Boost 3-d object detection via point
  clouds segmentation and fused 3-d giou-l1 loss,'' \emph{IEEE Transactions on
  Neural Networks and Learning Systems}, vol.~33, no.~2, pp. 762--773, 2022.

\bibitem{wang2021pointaugmenting}
C.~Wang, C.~Ma, M.~Zhu, and X.~Yang, ``Pointaugmenting: Cross-modal
  augmentation for 3d object detection,'' in \emph{Proceedings of the IEEE/CVF
  Conference on Computer Vision and Pattern Recognition}, 2021, pp.
  11\,794--11\,803.

\bibitem{multistage}
Z.~Wang, Z.~Zhao, Z.~Jin, Z.~Che, J.~Tang, C.~Shen, and Y.~Peng, ``Multi-stage
  fusion for multi-class 3d lidar detection,'' in \emph{2021 IEEE/CVF
  International Conference on Computer Vision Workshops (ICCVW)}, 2021, pp.
  3113--3121.

\bibitem{xie2019pircnn}
L.~Xie, C.~Xiang, Z.~Yu, G.~Xu, Z.~Yang, D.~Cai, and X.~He, ``Pi-rcnn: An
  efficient multi-sensor 3d object detector with point-based attentive
  cont-conv fusion module,'' 2019.

\bibitem{zhu2021crossmodal}
M.~Zhu, C.~Ma, P.~Ji, and X.~Yang, ``Cross-modality 3d object detection,'' in
  \emph{Proceedings of the IEEE/CVF Winter Conference on Applications of
  Computer Vision}, 2021, pp. 3772--3781.

\bibitem{chen2017multiviewmv3d}
X.~Chen, H.~Ma, J.~Wan, B.~Li, and T.~Xia, ``Multi-view 3d object detection
  network for autonomous driving,'' 2017.

\bibitem{ku2018jointAVOD}
J.~Ku, M.~Mozifian, J.~Lee, A.~Harakeh, and S.~Waslander, ``Joint 3d proposal
  generation and object detection from view aggregation,'' \emph{IROS}, 2018.

\bibitem{pang2020clocs}
S.~Pang, D.~Morris, and H.~Radha, ``Clocs: Camera-lidar object candidates
  fusion for 3d object detection,'' 2020.

\bibitem{Sun2020TowardsRL}
J.~Sun, Y.~Cao, Q.~A. Chen, and Z.~M. Mao, ``Towards robust lidar-based
  perception in autonomous driving: General black-box adversarial sensor attack
  and countermeasures,'' \emph{ArXiv}, vol. abs/2006.16974, 2020.

\bibitem{uscdmm}
K.~Huang, B.~Shi, X.~Li, X.~Li, S.~Huang, and Y.~Li, ``Multi-modal sensor
  fusion for auto driving perception: A survey,'' 2022.

\bibitem{xie2023robobev}
S.~Xie, L.~Kong, W.~Zhang, J.~Ren, L.~Pan, K.~Chen, and Z.~Liu, ``Robobev:
  Towards robust bird's eye view perception under corruptions,'' \emph{arXiv
  preprint arXiv:2304.06719}, 2023.

\bibitem{Yu2022BenchmarkingTRALI}
K.~Yu, T.~Tang, H.~Xie, Z.~Lin, Z.~Wu, Z.~Xia, T.~Liang, H.~Sun, J.~Deng,
  D.~Hao, Y.~Wang, X.~Liang, and B.~Wang, ``Benchmarking the robustness of
  lidar-camera fusion for 3d object detection,'' \emph{2023 IEEE/CVF Conference
  on Computer Vision and Pattern Recognition Workshops (CVPRW)}, pp.
  3188--3198, 2022.

\bibitem{bijelic2020seeingfog}
M.~Bijelic, T.~Gruber, F.~Mannan, F.~Kraus, W.~Ritter, K.~Dietmayer, and
  F.~Heide, ``Seeing through fog without seeing fog: Deep multimodal sensor
  fusion in unseen adverse weather,'' in \emph{Proceedings of the IEEE/CVF
  Conference on Computer Vision and Pattern Recognition}, 2020, pp.
  11\,682--11\,692.

\bibitem{Pitropov_2020cadc}
M.~Pitropov, D.~E. Garcia, J.~Rebello, M.~Smart, C.~Wang, K.~Czarnecki, and
  S.~Waslander, ``Canadian adverse driving conditions dataset,'' \emph{The
  International Journal of Robotics Research}, vol.~40, no. 4-5, pp. 681--690,
  12 2020.

\bibitem{DiazRuiz_2022_CVPRIthaca36}
C.~A. Diaz-Ruiz, Y.~Xia, Y.~You, J.~Nino, J.~Chen, J.~Monica, X.~Chen, K.~Luo,
  Y.~Wang, M.~Emond, W.-L. Chao, B.~Hariharan, K.~Q. Weinberger, and
  M.~Campbell, ``Ithaca365: Dataset and driving perception under repeated and
  challenging weather conditions,'' 2022.

\bibitem{HahnerCVPR22fogeth}
M.~Hahner, C.~Sakaridis, D.~Dai, and L.~Van~Gool, ``{Fog Simulation on Real
  LiDAR Point Clouds for 3D Object Detection in Adverse Weather},'' in
  \emph{IEEE International Conference on Computer Vision (ICCV)}, 2021.

\bibitem{HahnerCVPR22snow}
M.~Hahner, C.~Sakaridis, M.~Bijelic, F.~Heide, F.~Yu, D.~Dai, and L.~Van~Gool,
  ``{LiDAR Snowfall Simulation for Robust 3D Object Detection},'' in
  \emph{IEEE/CVF Conference on Computer Vision and Pattern Recognition (CVPR)},
  2022.

\bibitem{caesar2020nuScenes}
H.~Caesar, V.~Bankiti, A.~H. Lang, S.~Vora, V.~E. Liong, Q.~Xu, A.~Krishnan,
  Y.~Pan, G.~Baldan, and O.~Beijbom, ``nuscenes: A multimodal dataset for
  autonomous driving,'' 2020.

\end{thebibliography}
}

\end{document}